\documentclass{article}

 \usepackage[final]{nips_2018}

\usepackage[utf8]{inputenc} 
\usepackage[T1]{fontenc}    
\usepackage[draft]{hyperref}
\usepackage{url}            
\usepackage{booktabs}       
\usepackage{amsfonts}       
\usepackage{nicefrac}       
\usepackage{microtype}      
\usepackage{arydshln}
\usepackage{adjustbox}
\usepackage[justification=centering]{caption}
\usepackage{chngpage}
\usepackage{graphics}
\usepackage{floatrow}

\title{Machine Learning on Electronic Health Records: Models and Features Usages to predict Medication Non-Adherence}

\usepackage{authblk}

\author[, 1]{Janssoone T. \thanks{Corresponding author: \texttt{tjanssoone@semeia.io}}}
\author[1]{Bic C.}
\author[2]{Kanoun D..}
\author[1]{Rinder P.}
\author[1]{Hornus P.}
\affil[1]{Semeia, Paris, France}
\affil[2]{Institut Pasteur, Paris, France}

\begin{document}

\maketitle
\begin{abstract}
Adherence can be defined as \textit{"the extent to which patients take their medications as prescribed by their healthcare providers"}\citep{osterberg2005adherence}. World Health Organization's reports\footnote{\url{http://www.who.int/chp/knowledge/publications/adherence_full_report.pdf}} point out that, in developed countries, only about 50\% of patients with chronic diseases correctly follow their treatments. This severely compromises the efficiency of long-term therapy and increases the cost of health services.\\
We propose in this paper different models of patient drug consumption in breast cancer treatments. The aim of these different approaches is to predict medication non-adherence while giving insights to doctors of the underlying reasons of these illegitimate drop-outs. Working with oncologists, we show the interest of Machine-Learning algorithms fined tune by the feedback of experts to estimate a risk score of a patient's non-adherence and thus improve support throughout their care path.
\end{abstract}

\section{Introduction}
\citet{fallowfield2005patients, o2002oral} shows that patient-administered oral medications have become more and more prevalent, increasing the focus on adherence. A common solution is to set patient support programs that include, for example, 1) providing patients with information and advice 2) support and coaching sessions delivered by nurses (by phone or face-to-face), and 3) sending information to health professionals treating the patient. These programs have been shown to be effective \citep{krolop2013adherence, spoelstra2013intervention}. Yet, two main limitations are restraining these interventions: 1) The use of human intervention is effective but very expensive thus limiting their reach, 2) The use of digital technologies (e.g. notifications and explanations) is too generic and sometimes too intrusive (e.g. daily reminders) which leads to patients losing interest. We propose to optimize the use of human intervention by giving an accurate evaluation of a patient illegitimate drug drop-out. We use of machine-learning techniques on breast cancer patients' consumption data to evaluate this risk and fine-tune them with insights from oncologists. These predictive models are trained on the reimbursement data of the French Health Insurance (SNIIRAM - the French National Health System). The goal is to categorize patients into risk classes according to their characteristics. The long-term goal is to know the most appropriate moments to contact them for support. Thus, people will benefit from support adapted to their profile and their needs, and human interventions will be reserved for the situations for which they are really necessary. This paper first reviews previous approaches before introducing the SNIIIRAM database. Then, the different data processing and the applied models are described and discussed.

\section{Related work}

The phenomenon of non-adherence has been widely surveyed by medical researchers and psychologists. For instance, \citet{dimatteo2004variations} and \citet{mann2010predictors} review many observation-based scientific publications and provide an interesting quantitative assessment of the research conducted on the subject. These meta-analysis give us insights about factors of non-adherence. Hence, they underline the influence of the increasing age of patients, the treatment complexity level (multiple drugs, injections, \ldots) or the impact of patients' mental health ( depressive episodes have a very negative impact on the patient's compliance with the prescriptions of health professionals \citep{dimatteo2000depression}) as well as low education and, more importantly, low income \citep{dimatteo2004social}. However, \citet{franklin2016observing} underline the difficulty to use this information to predict adherence: they evaluate different approaches,  using logistic regression and boosted logistic regression, to define three categories of adherence predictors. Hence, they show that using census information or transaction data leads to poor prediction. However, they point out that using adherence observations during the first month significantly increases the accuracy of the results. This nuance on the weight of each adherence prediction variable is confirmed in \citet{lo2015using}. They use random survival forests highlights to find patient specific adherence thresholds to discriminate between hospitalization risks. Here again, the major variables are linked to patient history and previous transactions. \citet{karanasiou2016predicting} explore Machine Learning approaches to estimate the risk of the non-adherence for a patient with Heart Failure. The database contains general information about the patient, details about each care path, clinical examinations and clean labels based on clinician estimations. They prove the feasibility of this approach using a clean dedicated dataset. This paper explores these solutions to predict the risk of an illegitimate stop during a treatment. Our models are trained using real reimbursement records data from the French Health System. Our goal is to explore how to use these real, indirect and unlabeled observations to evaluate the risk of non-adherence at key moments of a patient's care path.


\section{The SNIIRAM database}

SNIIRAM\footnote{\url{https://www.ameli.fr/l-assurance-maladie/statistiques-et-publications/sniiram/finalites-du-sniiram.php}} is one of the largest structured databases of health data in the world. It contains reimbursement data of the French Health System, covering 99.8\% of the French population ($\simeq$ 66 million persons). Useful data are, for example, hospitalizations, drug purchases or contextual patient information (age, government services, geographic information, \ldots). The advantage of using this database is that it is very well structured and has no-bias in term of social background due to its universal coverage. More details can be found in \cite{tuppin2010french}. Previous work has already shown the value of massive data mining to aid diagnosis, either by taking all the information for a "static" approach \cite{neumann2012pioglitazone}, or, more recently, by also incorporating dynamic information \cite{morel2017convsccs}. Other studies have been conducted on the determinants of compliance, particularly for breast cancer. Our study focuses on women's breast cancer on part of the SNIIRAM data. The cohort of the study consists of 50$\%$ of women (drawn randomly) who met the following criteria: 1- diagnosed with breast cancer; 2- having purchased at least one of the following molecules for the studied period: \emph{Anastrozole}, \emph{Capecitabine}, \emph{Cyclophosphamide}, \emph{Etoposide}, \emph{Everolimus}, \emph{Exemestane}, \emph{Lapatinib}, \emph{Letrozole}, \emph{Megestrol}, \emph{Melphalan}, \emph{Tamoxifen}, \emph{Toremifene }and \emph{Vinorelbine}. Extraction concerns consumption between 2013 and 2015 and is made up of three main categories: 1-Pharmacy transactions (molecule, number of doses, date, \ldots); 2- Hospitalizations (diagnosis, start date, end date, \ldots); 3-Patient information (age, department, date of the diagnostic of eventual long-term illness (referred as \emph{ALD}) , pathologies, \ldots)

\section{Phase analysis study}
Working with oncologists, we reworked the raw data to show the different phases of the treatment. A phase is defined as a period of continuous intake of a molecule or hospitalization (chemotherapy or radiotherapy), allowing the reconstruction of a patient's care path. A phase of treatment can end with a legitimate stop (i.e. death, change of treatment, a serious cardiac issue, palliative care or right censorship due to data extraction) if this event occurs less than two months after the date of the last theoretical dose. The date of the last theoretical dose is obtained by calculating the median interval between two purchases of the molecule or two hospitalizations of the same type: this median behavior is considered to conform with the drug dose. Thus, the median time is 30 days between two box purchases of 30 doses of Tamoxifen. The end of this period after the last purchased box corresponds to the date of the last theoretical intake. Illegitimate stops are considered if none of the legitimate stops occurred before the date of the last theoretical dose. For each phase, the following data is calculated: 1- Start and end dates, number of intakes or hospitalizations, molecule or type of hospitalization; 2- End of treatment type (switch, death, stop, right censorship); 3-Patient information (comorbidities, location, age, \ldots); 4- Interventions on the breast (mastectomy) during the three months before the studied phase. This study focuses on predicting an illegitimate stop occurring after a specific time length (3, 6 and 12 months) from the beginning of the phase. For each phase, for each time period, a label indicates if an illegitimate stop occurred or if the care path continues correctly (no stop or a legitimate one). The data are then composed of, for the 3 months period, 51220 patients with 11.16$\%$ non-adherent, for 6 months, 44469 patients with 16.53$\%$ non-adherent, for 12 months, 34132 patients with 27.0$\%$ non-adherent.\\
We compare the ability of supervised learning algorithms to predict a legitimate stop from an illegitimate one with the information available at the beginning of the phase. The first model is a logistic regression, trained with a restricted number of variables according to their p-value ($<0.05$). Logistic regression assigns a coefficient to each variable related to its influence for the classification task, making it easy to interpret. The second model is a decision tree trained with all variables in the dataset. Parameters were tuned with gridsearch. Prediction function of a decision tree is easy to read and understand. The third model is gradient boosting which, compared to the two previous models, is harder to interpret to interpret. However, it focuses on weak learners and can offer a better prediction. This model used all variables in the dataset and parameters were tuned with grid search. The last model is a multilayer perceptron (MLP). Training sets are balanced to get an equal number of legitimate/illegitimate cases. We only use one hidden layer as balanced dataset have quite a small size (9412 rows for the 3 months period). We also kept all variables to compute the model. 5 folders cross-validation was performed which results are shown in Table\ref{tab:results_phases}. AUCs are roughly $0.70$ with Gradient Boosting having the best performances. A Cumulative Accuracy Profit curve (CAP or Lorenz curve) gives the second and third measures, indicating the ability of a model to accurately spot a patient at-risk. CAP $n\%$ is the rate of true positive classifications looking at the $n$ highest ranked predicted risk. Gradient boosting and MLP have the best results but these "black box methods" are difficult to explain which might refrain their usages. These first approaches could already be applied: the caregivers could be notified during the first medical visit and propose more support to the patient at risk. These risk models can also be used to underline the period with more drop-outs according to a patient's profile and trigger an alarm to reach them if needed.


\begin{table}[b!]
  \centering
  \begin{adjustbox}{max width=\textwidth}
\begin{tabular}[b]{|l||c|c|c||c|c|c||c|c|c|}
\hline
 & \multicolumn{3}{|c|}{3 months} & \multicolumn{3}{|c|}{6 months}  & \multicolumn{3}{|c|}{12 months}  \\
\hline
   & AUC & CAP 20$\%$ & CAP 40$\%$ & AUC & CAP 20$\%$ & CAP 40$\%$ & AUC & CAP 20$\%$ & CAP 40$\%$\\
\hline \hline
Logistic regression   & 0.71 & 0.52 & 0.72 & 0.71 & 0.52 & 0.71 & 0.72 & 0.54 & 0.71 \\
Decision tree         & 0.68 & 0.47 & 0.68 & 0.69 & 0.47 & 0.68 & 0.70 & 0.50 & 0.69 \\
Gradient boosting     & 0.74 & 0.54 & 0.75 & 0.74 & 0.55 & 0.74 & 0.74 & 0.57 & 0.75 \\
Multilayer perceptron & 0.71 & 0.52 & 0.72 & 0.71 & 0.53 & 0.72 & 0.72 & 0.55 & 0.71 \\
\hline
\end{tabular}
\end{adjustbox}
  \caption{AUCs and CAP results of the algorithms ability to predict an illegitimate stops in the n months after the beginning of a phase}
  \label{tab:results_phases}
\end{table}

\section{Drug transaction study}

A survival analysis, described in \citet{janssoone2018predictive}, and a statistical analysis of the pharmacy transactions show that illegitimate drop out rate evolves over time and according to previous events in the patient's care path. This led to a more patient-centered study focusing on all transactions carried out by a patient: every purchase in a pharmacy, every hospitalization,\ldots For each pharmacy transaction of a drug used for treating breast cancer (Tamoxifen, Exemestane,\ldots), the model predicts if the patient will continue her treatment correctly or if an illegitimate stop might occur. Same criteria used in the phases study are used to label pharmacy transactions. Data can, therefore, be considered as a sequence of events (hospitalizations and pharmacy transactions) with timestamps and a set of characteristics about the patient. We can divide data in two categories : \emph{"static"} data describing the patient (geographic and sociological information, age, \ldots), and \emph{dynamic} data about the current care-path( last drug purchases (cancer linked and general ones), co-pathologies, \ldots). The latter can be seen as a sequence of events on which Recurrent Neural Networks (RNN) are known to perform well, especially \emph{Long short-term memory} (LSTM \citet{hochreiter1997long}).\\
Our ongoing study obtains its best results with the following architecture: 1- an LSTM network is applied on the "dynamic" information (pharmacy transactions and hospitalizations); 2- an MLP network on the "contextual" information (details about the patient (geographical, financial support, \ldots)) which are not frequently updated in the SNIIRAM database; 3- both outputs are concatenated and then classified through a fully-connected layer. We tested networks based on this kind of architecture. "Dynamic" inputs are sequences of 10 observations and the output indicates if the current transaction might not be followed by another one (indicating the risk of an illegitimate drop-out). As 10 observations are not always available for each transaction, we also test different padding preprocessing with zeros filling or first observation duplication. In this study, zeros filling got the best results and was retained for our ongoing process. The underlying goal of this study is also to gain insights into the influence of different sets of features. As it is doable to analyze the use of static information through the MLP network, we focus here on the dynamic data as RNN are more tricky to understand. Our strategy is to train different network with more or fewer features data. Our baseline just handles cancerology linked pharmacy transaction and longtime hospitalizations. Then we add co-pathologies and other pharmacy transactions. Yet, more dynamic information does not increase the performance as we always obtain a score that allows targeting efficiently a patient in 82\% of the cases. The CAP curve indicates that, within the first $20\%$ of the surveyed population ranked from the highest to the lowest risk estimated, our models target 66\% of illegitimate stops of all transactions. This is three times as effective than the current random model used to target patients at risk, thus validating our hypothesis that last transactions convey information to detect an illegitimate stop.\\ 
As French Health Services have a limited number of hours to call patients for support, our model could double their efficiency. This results could also be applied to notify pharmacists to deliver more support when appropriate or trigger an SMS based system to contact and motivate a high-risk patient. Yet, the different data-usage we tried don't show a lot of improvements in our results. We might use deeper networks to gain a few points in AUC and CAP curves. Our first attempts to do so resulted in too much instability in our training looking at the standard deviation over the k-fold. This could be explained by the need for more data to provide stable results for more complex networks. We plan to test this hypothesis but it might take some times due to the process of SNIIRAM data extraction (which is very regarding the respect of privacy of the patients). We also look at other solutions such as using Generative Adversarial Network to increase the size of data or perform transfer-learning using other Electronic Health Records to improve the general information about care paths of patients.

\section{Conclusion and discussion}
This paper presents our first studies on how machine-learning can be used to help patients follow their treatment during long-term illness. We explore several approaches applied to the French Health System's reimbursement data to estimate the risk of an illegitimate drug drop-out. The results obtained with simple models on indirect observations from SNIIRAM (reimbursement data) prove the feasibility of our process. We validate our first results with feedbacks from oncologists and medical researchers. Our approaches also aim to be coherent with patients' care-paths. Using drug-phases formalism or looking directly at pharmacy-transactions, we show that we could notify caregivers of the potential risk of a drug drop-out at specific moments of the treatment. This allows to provide a more efficient support at appropriate times while avoiding stress resulting from too frequent unnecessary contacts and limiting the waste of resources for low-risk patients.\\
Both studies show their abilities to predict patients'non-adherence. One main limitation is the intelligibility of their decision process which remains hard to interpret with "black box methods". It will be a challenge as medical staffs want to understand the path of patients. Yet, we validate the ability of AI to estimate the risk of non-adherence. More complex networks should improve our efficiency. For example, we could look at several pathologies and use some kind of domain adaptation methods to detect patterns relevant to non-adherence. The other main challenge concerns the labeling of the data: we plan to explore automatic labeling and anomaly discovery to find more accurate information in our data.

\newpage
\bibliographystyle{plainnat}
\bibliography{biblio}

\begin{thebibliography}{17}
\providecommand{\natexlab}[1]{#1}
\providecommand{\url}[1]{\texttt{#1}}
\expandafter\ifx\csname urlstyle\endcsname\relax
  \providecommand{\doi}[1]{doi: #1}\else
  \providecommand{\doi}{doi: \begingroup \urlstyle{rm}\Url}\fi

\bibitem[DiMatteo(2004{\natexlab{a}})]{dimatteo2004social}
M~Robin DiMatteo.
\newblock Social support and patient adherence to medical treatment: a
  meta-analysis.
\newblock \emph{Health psychology}, 23\penalty0 (2):\penalty0 207,
  2004{\natexlab{a}}.

\bibitem[DiMatteo(2004{\natexlab{b}})]{dimatteo2004variations}
M~Robin DiMatteo.
\newblock Variations in patients’ adherence to medical recommendations: a
  quantitative review of 50 years of research.
\newblock \emph{Medical care}, 42\penalty0 (3):\penalty0 200--209,
  2004{\natexlab{b}}.

\bibitem[DiMatteo et~al.(2000)DiMatteo, Lepper, and
  Croghan]{dimatteo2000depression}
M~Robin DiMatteo, Heidi~S Lepper, and Thomas~W Croghan.
\newblock Depression is a risk factor for noncompliance with medical treatment:
  meta-analysis of the effects of anxiety and depression on patient adherence.
\newblock \emph{Archives of internal medicine}, 160\penalty0 (14):\penalty0
  2101--2107, 2000.

\bibitem[Fallowfield et~al.(2005)Fallowfield, Atkins, Catt, Cox, Coxon,
  Langridge, Morris, and Price]{fallowfield2005patients}
L~Fallowfield, L~Atkins, S~Catt, A~Cox, C~Coxon, C~Langridge, R~Morris, and
  M~Price.
\newblock Patients' preference for administration of endocrine treatments by
  injection or tablets: results from a study of women with breast cancer.
\newblock \emph{Annals of Oncology}, 17\penalty0 (2):\penalty0 205--210, 2005.

\bibitem[Franklin et~al.(2016)Franklin, Shrank, Lii, Krumme, Matlin, Brennan,
  and Choudhry]{franklin2016observing}
Jessica~M Franklin, William~H Shrank, Joyce Lii, Alexis~K Krumme, Olga~S
  Matlin, Troyen~A Brennan, and Niteesh~K Choudhry.
\newblock Observing versus predicting: Initial patterns of filling predict
  long-term adherence more accurately than high-dimensional modeling
  techniques.
\newblock \emph{Health services research}, 51\penalty0 (1):\penalty0 220--239,
  2016.

\bibitem[Hochreiter and Schmidhuber(1997)]{hochreiter1997long}
S.~Hochreiter and J.~Schmidhuber.
\newblock Long short-term memory.
\newblock \emph{Journal of Neural computation}, 1997.

\bibitem[Janssoone et~al.(2018)Janssoone, Rinder, Hornus, and
  Kanoun]{janssoone2018predictive}
T~Janssoone, P~Rinder, P~Hornus, and D~Kanoun.
\newblock Predictive patient care: Survival model to prevent medication
  non-adherence.
\newblock 2018.

\bibitem[Karanasiou et~al.(2016)Karanasiou, Tripoliti, Papadopoulos, Kalatzis,
  Goletsis, Naka, Bechlioulis, Errachid, and
  Fotiadis]{karanasiou2016predicting}
Georgia~Spiridon Karanasiou, Evanthia~Eleftherios Tripoliti,
  Theofilos~Grigorios Papadopoulos, Fanis~Georgios Kalatzis, Yorgos Goletsis,
  Katerina~Kyriakos Naka, Aris Bechlioulis, Abdelhamid Errachid, and
  Dimitrios~Ioannis Fotiadis.
\newblock Predicting adherence of patients with hf through machine learning
  techniques.
\newblock \emph{Healthcare technology letters}, 3\penalty0 (3):\penalty0
  165--170, 2016.

\bibitem[Krolop et~al.(2013)Krolop, Ko, Schwindt, Schumacher, Fimmers, and
  Jaehde]{krolop2013adherence}
Linda Krolop, Yon-Dschun Ko, Peter~Florian Schwindt, Claudia Schumacher, Rolf
  Fimmers, and Ulrich Jaehde.
\newblock Adherence management for patients with cancer taking capecitabine: a
  prospective two-arm cohort study.
\newblock \emph{BMJ open}, 3\penalty0 (7):\penalty0 e003139, 2013.

\bibitem[Lo-Ciganic et~al.(2015)Lo-Ciganic, Donohue, Thorpe, Perera, Thorpe,
  Marcum, and Gellad]{lo2015using}
Wei-Hsuan Lo-Ciganic, Julie~M Donohue, Joshua~M Thorpe, Subashan Perera,
  Carolyn~T Thorpe, Zachary~A Marcum, and Walid~F Gellad.
\newblock Using machine learning to examine medication adherence thresholds and
  risk of hospitalization.
\newblock \emph{Medical care}, 53\penalty0 (8):\penalty0 720, 2015.

\bibitem[Mann et~al.(2010)Mann, Woodward, Muntner, Falzon, and
  Kronish]{mann2010predictors}
Devin~M Mann, Mark Woodward, Paul Muntner, Louise Falzon, and Ian Kronish.
\newblock Predictors of nonadherence to statins: a systematic review and
  meta-analysis.
\newblock \emph{Annals of Pharmacotherapy}, 44\penalty0 (9):\penalty0
  1410--1421, 2010.

\bibitem[Morel et~al.(2017)Morel, Bacry, Ga{\"\i}ffas, Guilloux, and
  Leroy]{morel2017convsccs}
Maryan Morel, Emmanuel Bacry, St{\'e}phane Ga{\"\i}ffas, Agathe Guilloux, and
  Fanny Leroy.
\newblock Convsccs: convolutional self-controlled case series model for lagged
  adverse event detection.
\newblock \emph{arXiv preprint arXiv:1712.08243}, 2017.

\bibitem[Neumann et~al.(2012)Neumann, Weill, Ricordeau, Fagot, Alla, and
  Allemand]{neumann2012pioglitazone}
A~Neumann, A~Weill, P~Ricordeau, JP~Fagot, F~Alla, and H~Allemand.
\newblock Pioglitazone and risk of bladder cancer among diabetic patients in
  france: a population-based cohort study.
\newblock \emph{Diabetologia}, 55\penalty0 (7):\penalty0 1953--1962, 2012.

\bibitem[O'neill and Twelves(2002)]{o2002oral}
VJ~O'neill and CJ~Twelves.
\newblock Oral cancer treatment: developments in chemotherapy and beyond.
\newblock \emph{British journal of cancer}, 87\penalty0 (9):\penalty0 933,
  2002.

\bibitem[Osterberg and Blaschke(2005)]{osterberg2005adherence}
L~Osterberg and T~Blaschke.
\newblock Adherence to medication.
\newblock \emph{New England Journal of Medicine}, 353\penalty0 (5):\penalty0
  487--497, 2005.

\bibitem[Spoelstra et~al.(2013)Spoelstra, Given, Given, Grant, Sikorskii, You,
  and Decker]{spoelstra2013intervention}
Sandra~L Spoelstra, Barbara~A Given, Charles~W Given, Marcia Grant, Alla
  Sikorskii, Mei You, and Veronica Decker.
\newblock An intervention to improve adherence and management of symptoms for
  patients prescribed oral chemotherapy agents: an exploratory study.
\newblock \emph{Cancer nursing}, 36\penalty0 (1):\penalty0 18--28, 2013.

\bibitem[Tuppin et~al.(2010)Tuppin, De~Roquefeuil, Weill, Ricordeau, and
  Merli{\`e}re]{tuppin2010french}
P~Tuppin, L~De~Roquefeuil, A~Weill, P~Ricordeau, and Y~Merli{\`e}re.
\newblock French national health insurance information system and the permanent
  beneficiaries sample.
\newblock \emph{Revue d'epidemiologie et de sante publique}, 58\penalty0
  (4):\penalty0 286--290, 2010.

\end{thebibliography}
\end{document}